# Methodological Explainability Evaluation of an Interpretable Deep Learning Model for Post-Hepatectomy Liver Failure Prediction Incorporating Counterfactual Explanations and Layerwise Relevance Propagation: A Prospective *In Silico* Trial

Xian Zhong[1,3,*], Zohaib Salahuddin[1,*], Yi Chen[1,4], Henry C Woodruff[1,2], Haiyi Long[3], Jianyun Peng[3], Nuwan Udawatte[1], Roberto Casale[5], Ayoub Mokhtari[5], Xiaoer Zhang[3], Jiayao Huang[3], Qingyu Wu[3], Li Tan[6], Lili Chen[7], Dongming Li[6], Xiaoyan Xie[3], Manxia Lin[3, †], Philippe Lambin[1,2, †]

1.  The D-Lab, Department of Precision Medicine, GROW – School for Oncology, Maastricht University, Maastricht, the Netherlands
2.  Department of Radiology and Nuclear Medicine, GROW – School for Oncology, Maastricht University Medical Center+, Maastricht, the Netherlands
3.  Department of Medical Ultrasonics, Institute of Diagnostic and Interventional Ultrasound, The First Affiliated Hospital of Sun Yat-sen University, Guangzhou, China
4.  Key Laboratory of Intelligent Medical Image Analysis and Precise Diagnosis, College of Computer Science and Technology, Guizhou University, Guiyang, China
5.  Radiology Department, Institut Jules Bordet Hôpital Universitaire de Bruxelles, Université Libre de Bruxelles, 1070 Brussels, Belgium
6.  Center of Hepato-Pancreato-Biliary Surgery, The First Affiliated Hospital of Sun Yat-sen University, Guangzhou, China
7.  Department of Pathology, The First Affiliated Hospital of Sun Yat-sen University, Guangzhou, China
8.  Center of Hepato-Pancreato-Biliary Surgery, The First Affiliated Hospital of Sun Yat-sen University, Guangzhou, China

## Abstract

Artificial intelligence (AI)-based decision support systems have demonstrated value in predicting post-hepatectomy liver failure (PHLF) in hepatocellular carcinoma (HCC). However, they often lack transparency, and the impact of model explanations on clinicians' decisions has not been thoroughly evaluated. Building on prior research, we developed a variational autoencoder-multilayer perceptron (VAE-MLP) model for preoperative PHLF prediction. This model integrated counterfactuals and layerwise relevance propagation (LRP) to provide insights into its decision-making mechanism. Additionally, we proposed a methodological framework for evaluating the explainability of AI systems. This framework includes qualitative and quantitative assessments of explanations against recognized biomarkers, usability evaluations, and an *in silico* clinical trial. Our evaluations demonstrated that the model's explanation correlated with established biomarkers and exhibited high usability at both the case and system levels. Furthermore, results from the three-track *in silico* clinical trial showed that clinicians' prediction accuracy and confidence increased when AI explanations were provided.

---

[*] These authors contributed equally as first authors
[†] These authors contributed equally as senior authors

## 1. Introduction

Hepatocellular carcinoma (HCC) is the sixth most common malignancy and the third leading cause of cancer-related death globally[1]. Liver resection is the main curative treatment modality for HCC[2]. Despite the advancement of surgical techniques, post-hepatectomy liver failure (PHLF) remains the major cause of postoperative morbidity and mortality after liver resection in HCC, with an overall incidence of up to 32% and corresponding mortality of up to 5.0%[3]. Preoperative prediction of PHLF is of great importance to improve perioperative management and optimize treatment options such as switching to alternative treatments like radiofrequency ablation (RFA) or transarterial chemoembolization (TACE)[4]. This underscores the critical need to explore a decision support system for PHLF prediction. Some clinical variables such as future liver remnant (FLR) volume, Child-Pugh score, Model for end-stage liver disease (MELD), and Albumin-Bilirubin (ALBI) have been reported for PHLF prediction[5,6], however, all of them had limited prediction accuracy. Two-dimensional shear wave elastography (2D-SWE) is a liver stiffness measurement (LSM) technology that has been proven to be useful in liver fibrosis staging[7,8]. Therefore 2D-SWE shows potential for liver function assessment and PHLF prediction[9,10]. However, routine 2D-SWE analysis often underutilizes available image information and faces challenges due to inter-observer variance in quantification region selection[11]. Recent advances in artificial intelligence (AI)-based decision support systems have improved the analysis of 2D-SWE images, offering enhanced diagnostic capabilities[12,13].

AI and deep learning (DL) have demonstrated state-of-the-art performance on many medical imaging tasks such as diagnosis or prediction. However, despite significant progress, the clinical translation of DL tools has so far been limited, partially due to a lack of interpretability of models, the so-called "black box" problem[14]. The terms explainability and interpretability are often used interchangeably and refer to any attempt to understand the underlying decision-making processes of deep neural networks[15]. Interpretability of DL systems is important for fostering clinical trust as well as the timely correction of any faulty processes in the algorithm. Moreover, it is also one of the requirements of the European General Data Protection Regulation (GDPR)[16]. Many interpretability methods have been used for understanding deep learning models. These interpretability methods can be classified into various types, such as concept learning models[17], attribution maps[18], and concept attribution[19], among others. Most of these methods concentrate on a single aspect of explanation, such as images. However, many clinical decision are made not only based on the imaging information, but also the clinical information[20,21]. Integrating multiple explainability approaches can provide complementary information, particularly for methods that incorporate various streams of data, such as images and clinical variables.

Attribution maps are the most common explainability method for deep neural networks, highlighting key regions important for predictions[22,23]. However, these methods merely indicate the areas of interest without demonstrating how variations within these regions affect model performance[24]. Counterfactual explanations allow users to explore "what-if" scenarios, which helps in identifying areas of greatest importance, as well as aiding in understanding the changes that need to be made to switch the classifier's prediction[25,26]. Recent progress in generative models such as variational autoencoder (VAE) has led to the ability to provide insights into model behavior by generating new images under different conditions[27,28]. These counterfactual images are generated by applying minimal perturbations to the original image to achieve a maximum change in the classifier's prediction thus altering the predicted class of the original image[29]. It's also important to determine the contribution of each input feature to the prediction for explanation and validation of the model. The LRP method generates explanations by computing the relevance score of each neuron within a specific layer using propagation rules by moving in reverse from the output to the input[30,31]. The relevance score represents the quantitative contribution of a given neuron to the prediction. In various medical applications, LRP was shown to produce reliable explanations[32,33].

The evaluation of the reliability and effectiveness of explanations is crucial before they can be utilized in a clinical setting to determine the impact of the model explanations on diagnostic or predictive accuracy[14]. Early research mainly focused on qualitative or quantitative validation of explanations against established biomarkers and existing clinical knowledge to validate technical correctness. This evaluation included human perception[34], qualitative visualizations[26] or quantitative metrics[35], while evaluation with end-users was highly uncommon[36]. However, since the application of interpretable AI needs human-machine interaction, it's important to consider the effects of model explanation on users' experience with the system and their ability to act on the model's outputs[37]. Therefore, human-factors engineering items such as understandability, reliability, trust, or satisfaction should be also evaluated. The System Causability Scale (SCS), which is derived from the System Usability Scale (SUS), serves to evaluate the effectiveness of these explanations by measuring their utility within the collaborative clinician-AI explanation interface[38]. Furthermore, the influence of model explanations on diagnostic or predictive accuracy in a collaborative setting remains uncertain. For example, a study on diabetic retinopathy grading by ten ophthalmologists demonstrated that integrated gradient-based heatmap explanations resulted in overdiagnosis [39]. Conversely, another study showed that dermatologists' diagnostic accuracy, confidence in their diagnoses, and trust in the support system significantly increased with explainable AI compared to conventional AI[40]. Therefore, it is important to investigate the impact of introducing AI explanations in a collaborative setting involving clinicians. *In silico* trials (IST) allow the use of retrospective data in a prospective fashion by offering tight control over the ability to simulate various scenarios in an efficient and cost-effective manner [41,42]. An IST tool can be utilized to quantitatively assess the impact of AI tool's recommendations in a collaborative setting[43].

Here, we presented a novel interpretable deep learning model for PHLF prediction which incorporates counterfactual analysis and LRP for the explanation of both medical images and clinical variables. A methodological framework was proposed for the evaluation of model explainability. The key contributions of the work were as follows:

- We proposed a variational autoencoder-multilayer perceptron (VAE-MLP) model that used 2D-SWE images and clinical variables to predict PHLF in HCC. The proposed model integrates two complementary streams of explanation, namely counterfactuals, and LRP, to offer insights into the decision-making mechanism of the model, providing model explanation for both medical images and clinical variables. The proposed VAE-MLP model can be generalized to other medical AI applications with a better model explanation.
- We are one of the first studies to propose a methodological framework for the explainability evaluation of deep neural networks for clinical use, as shown in Figure 1, which includes qualitative and quantitative assessments of explanations against recognized biomarkers, usability evaluations via human-grounded studies, and a multi-institutional *in silico* clinical trial that includes efficacy evaluations in a three-track manner: without AI assistance, with AI assistance, and with AI plus explainability assistance.
- A usability evaluation of explanations was conducted with clinicians from multiple centers on various aspects of counterfactual and LRP explanations, focusing on their understandability, the justification of the classifier's decisions, helpfulness, and confidence in the explanations provided. Additionally, the overall quality of the explanations in the proposed model was assessed using the System Causability Scale.
- Through a three-track *in silico* clinical trial, we assessed the impact of model explanations on clinicians' prediction accuracy, confidence, and explanation satisfaction. Our results highlighted the significant impact of AI assistance and model explanation on clinicians' prediction accuracy and confidence with high satisfaction.

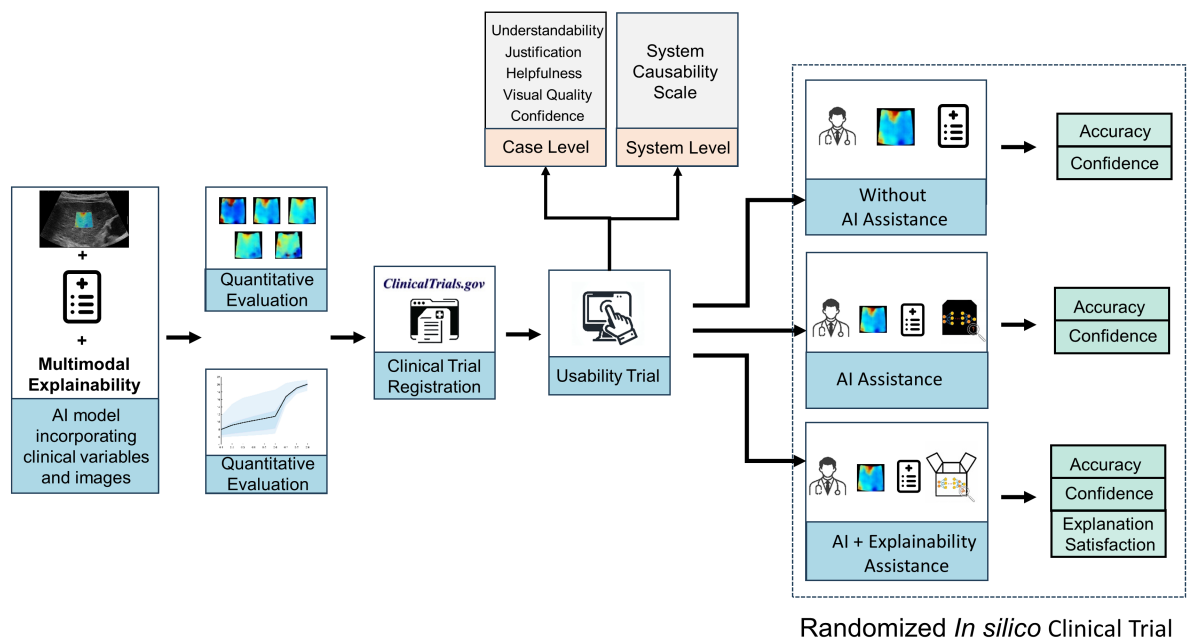

**Fig. 1** Proposed framework for evaluating explainable AI algorithms in healthcare. The flowchart outlines the multimodal assessment process for AI explainability, starting with qualitative and quantitative evaluation for the explainability of AI models. The framework progresses through clinical trial registration, a usability trial, and concludes with an *in silico* clinical trial conducted in three tracks: without AI assistance, with AI assistance, and with AI plus explainability assistance.

## 2. Materials and Methods

### 2.1 Dataset and data collection

The study protocol received approval from the Institutional Review Board of the First Affiliated Hospital of Sun Yat-sen University in China (IRB approval number: [2019]046). Prior to enrollment, all patients provided written informed consent.

Patients eligible for curative liver resection for HCC between August 2018 and October 2022 at the First Affiliated Hospital of Sun Yat-sen University in China were included. A total of 345 patients were included, comprising 265 patients in the training set and 80 patients in the test set. More detailed information on the inclusion and exclusion criteria, as well as the 2D-SWE and clinical information collection, can be found in Supplementary Method A1.

### 2.2 Diagnosis and staging of symptomatic PHLF

PHLF was defined as an increased international normalized ratio (INR) and hyperbilirubinemia on or after postoperative day 5, as proposed by the International Study Group of Liver Surgery (ISGLS)[3]. The grade of PHLF is categorized based on its impact on clinical management. Grade A PHLF indicates no alterations in treatment protocol, while grade B PHLF necessitates modifications in the standard treatment regimen without invasive interventions. In contrast, grade C PHLF requires invasive therapeutic measures. The symptomatic PHLF group was defined as having PHLF grade B or greater, whereas those with grade A PHLF or without PHLF symptoms fall into the non-symptomatic PHLF category[44].

## 2.3 Image preprocessing

The 2D-SWE bounding box was automatically generated from DICOM images, comprising both elastographic and B-mode images. The color elasticity image was derived by subtracting 50% of the corresponding B-mode image from the composite image, which was then resized to 128×128 pixels. Subsequently, the circular measurement markers in the 2D-SWE images (Q-box) were identified, and each pixel was substituted with the average value of the adjacent 4×4 pixel area.

## 2.4 $\beta$-Variational Autoencoder

VAEs integrate Bayesian variational inference in the autoencoder to introduce continuity in the latent space, making VAEs generative models [45]. VAEs consist of two main building blocks, an encoder $E_\theta(x_i)$ and a decoder $D_\varphi(z_i)$. Let $X = \{x_1, x_2, \ldots, x_n\}$ be n training images then the encoder $E_\theta(x_i)$ encodes the image $x_i$ to a latent representation $z_i$. The decoder $D_\theta(z_i)$ utilises the encoded latent space representation $Z = \{z_1, z_2, \ldots, z_n\}$ to reconstruct the input image $\underline{X} = \{\underline{x_1}, \underline{x_2}, \ldots, \underline{x_n}\}$. In contrast to simple autoencoders, VAEs learn to represent each latent attribute as a probability distribution. The loss used to train the VAE consists of a reconstruction loss that ensures that the reconstructed image from the latent space is similar to the input image, and a Kullback-Leiber (KL) divergence loss that ensures that the learned distribution is similar to the prior distribution. The loss function used to train the VAE is given by:

$$L(\theta, \varphi; x, z) = E_{q_\varphi(z|x)}[log\, p_\theta(x|z)] - D_{KL}(q_\varphi(z|x) \,||\, p(z))$$

where $E_{q_\varphi(z|x)}[log\, p_\theta(x|z)]$ was the reconstruction likelihood while $D_{KL}(q_\varphi(z|x) \,||\, p(z))$ is the KL divergence. $\beta$-VAE introduces a parameter $\beta > 1$ in the regularisation term to improve the qualitative nature of the disentangled representations learned by model [46]. The loss function to train $\beta$-VAE is given by:

$$L(\theta, \varphi, \beta; x, z) = E_{q_\varphi(z|x)}[log\, p_\theta(x|z)] - \beta \cdot D_{KL}(q_\varphi(z|x) \,||\, p(z))$$

We used the MONAI implementation of the variational autoencoder [47]. Scaling (0.9 times, 1.1 times), shifting (0 percent, 10 percent), rotation (-20 degrees, 20 degrees), and horizontal and vertical flipping (probability =0.5) augmentation were applied during training. The input channels were configured with three channels: red, green, and blue (RGB). The size of the latent space was set to 256. The number of channels at each layer of the encoder was set to (16, 32, 64, 128, 256) and the strides were set to (1, 2, 2, 2, 2). The $\beta$ variable in the loss function was set to 2.5. We used Adam optimizer with a learning rate of $10^{-5}$, and a ReducedLROnPlateau learning rate scheduler. $\beta$-VAE was trained for 250 epochs.

The $\beta$-VAE converted the 2D-SWE images into 256 latent space encodings. The latent space of the $\beta$-VAE was used to train a multilayer perceptron model ($VAE - MLP_{swe}$) consisting of 4 linear layers. Each linear layer was followed by a batch normalization and a leakyReLU layer with a negative slope=0.2. The MLP was trained with an Adam optimizer with a learning rate of $10^{-4}$, weight decay of 0.1, and a ReduceLROnPlateau learning rate scheduler. The MLP was trained for 100 epochs.

We incorporated all clinical variables with p<0.1 in univariate logistic analysis into the latent space of the VAE to classify PHLF using both the clinical variables and the 2D-SWE images. $VAE - MLP_{swe-cl}$ was a modification of $VAE - MLP_{swe}$ such that selected clinical variables were incorporated in the latent space of $VAE - MLP_{swe}$ (Fig. 2a).

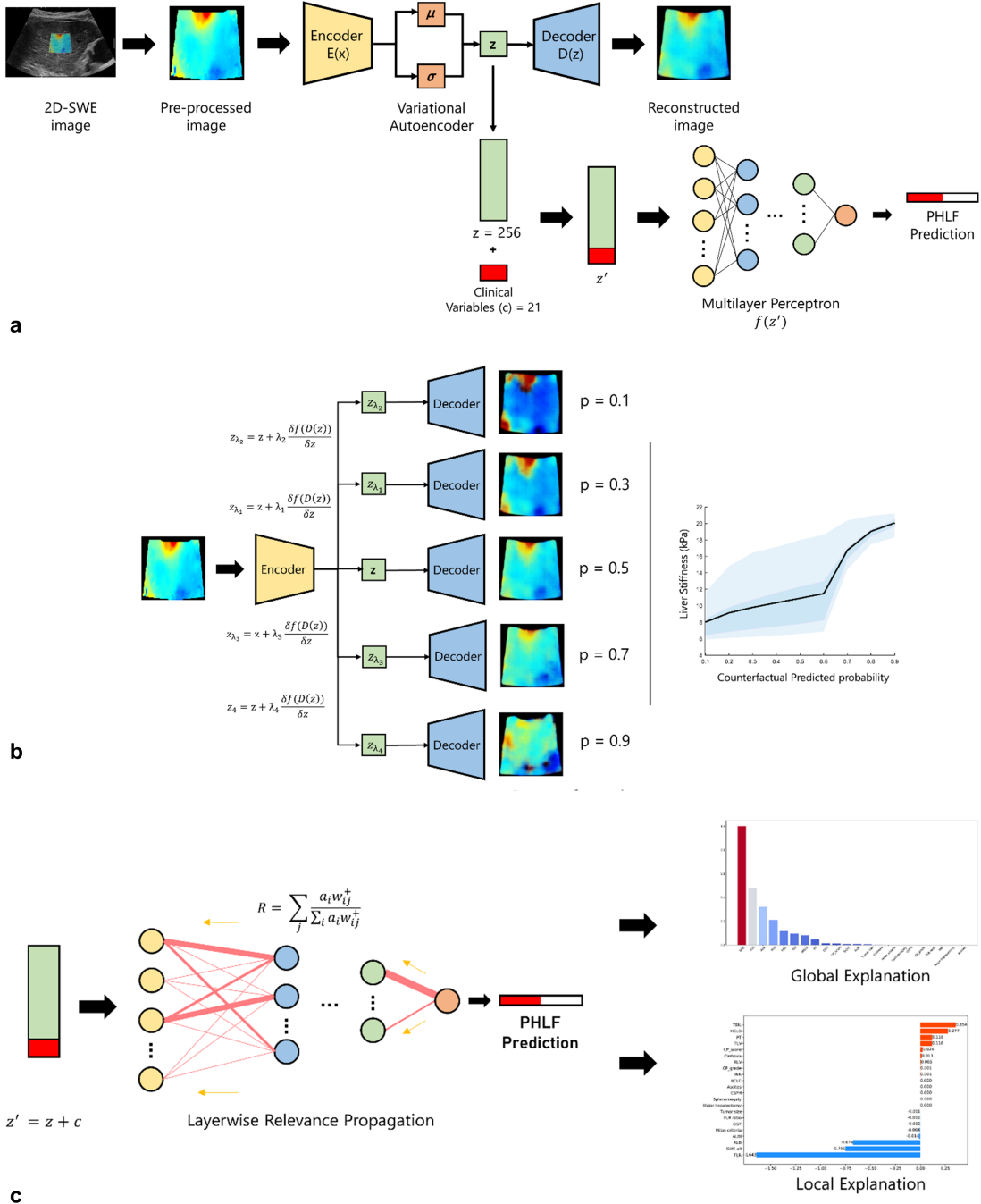

**Fig. 2** The proposed model for PHLF prediction incorporates counterfactual and layerwise relevance methods for explainability. (a) The workflow of the VAE-MLP model. (b) The workflow of counterfactual explanations for qualitative and quantitative analysis. (c) The workflow of layerwise relevance propagation for local and global explanations.

## 2.5 Counterfactual explanation

Counterfactual explanations were generated by applying minimal perturbation to the input image in such a manner that the prediction of the classifier was switched [14]. We trained an MLP using the latent space representations of $\beta$-VAE as an input. We can exploit the continuity in the latent space of $\beta$-VAE to perturb the latent space and generate new images using the decoder. We followed the methodology in [48] to apply the modification to the semantically important pixels for classification using the gradient of the classifier. Let $f$ denote the MLP classifier. The encoder $E_\theta(x)$ encoded the input image $x$ and generated a latent space representation $z$. The decoder $D_\varphi(z)$ converted the latent representation $z$ to a reconstructed image $\bar{x}$. Latent space encoding $z$ was perturbed in the direction of the gradient $\frac{\delta f(z)}{\delta z}$ of the classifier $f$ by an amount $\lambda$. The perturbation in the latent space was applied as follows:

$$z_\lambda = z + \lambda \cdot \frac{\delta f(z)}{\delta z}$$

The perturbed latent space representation $z_\lambda$ was fed to a decoder $D_\varphi(z_\lambda)$ to produce a new image $\bar{x}_\lambda$.

$$\bar{x}_\lambda = D_\varphi(z_\lambda)$$

By progressively perturbing the latent space to generate a higher or lower prediction $f(z_\lambda)$, we could observe the meaningful semantic features that the classifier considered important for prediction.

### 2.6 Qualitative and quantitative clinical relevance evaluation of counterfactual explanations

We generated the counterfactual images with predicted probabilities of 0.1-0.9 with an interval of 0.1 for all test cases in the $VAE - MLP_{swe-cl}$ model. For qualitative evaluation, we assessed the main visual changes of the counterfactual-generated images along with the changes in predicted probabilities of the counterfactual-generated images. For quantitative evaluation, we demonstrated the clinical relevance of counterfactual explanations using liver stiffness derived from these images. For each counterfactual-generated image, a circular region of interest (ROI) of 64-pixel diameter was placed on the most homogeneous area automatically to derive liver stiffness. The average value of liver stiffness of all test images with a predicted probability of 0.1-0.9 was calculated. The relationship between mean liver stiffness and predicted probability of counterfactual generated images was evaluated (Fig. 2b).

### 2.7 Layerwise Relevance Propagation (LRP)

LRP is an attribution method that calculates the contribution of each neuron by propagating the prediction backward based on relevance scores [30]. The total relevance at each layer of the neural network remains the same starting from the last layer of the classifier $f$. LRP calculates the magnitude of the relevance of each neuron by propagating the contribution starting from the last layer and moving backward one layer at a time. Let $R_a^{(l)}$ represent the relevance score of a neuron $a$ in layer $l$ and $R_b^{(l+1)}$ represent the relevance score of a neuron $b$ in layer $l + 1$. The relevance score $R_a^{(l)}$ is calculated by propagating the relevance scores of neurons in layer $l + 1$ as follows:

$$R_a^{(l)} = \Sigma_b \frac{a_a^{(l)} \cdot w_{ab}}{\Sigma_a a_a^{(l)} \cdot w_{ab}}$$

where $a_a^{(l)}$ is the activation of neuron $a$ in layer $l$ and $w_{ab}$ is the weight between neuron $a$ in layer $l$ and neuron $b$ in layer $l + 1$.

We utilized LRP for interpretability in two different ways (Fig. 2c):

1. Local explanation: We leveraged the principle of LRP that the total relevance at each layer of the neural network remains the same. A local LRP plot was generated to investigate the contribution of 2D-SWE and clinical variables in $VAE - MLP_{swe-cl}$ for prediction in each individual test case.

2. Global explanation: A global LRP plot was generated by aggregating the contribution of 2D-SWE and clinical variables over the entire test set. This showed how important each variable was for $VAE - MLP_{swe-cl}$ for classification in all test cases.

**2.8 A web-based *in silico* trial platform**

A web-based *in silico* trial platform ([https://st-trial-hbhdzvtdqlr.streamlit.app/](https://st-trial-hbhdzvtdqlr.streamlit.app/)) was established for both usability trials and *in silico* clinical trials. The platform was built based on Streamlit ([https://streamlit.io/](https://streamlit.io/)), an easy-to-use tool that turns data scripts into shareable web apps. The results were automatically recorded through Cloud Firestore (https://firebase.google.com/products/firestore/), a flexible, scalable NoSQL cloud database built on Google Cloud infrastructure. Before participating in the trials, every participant underwent a comprehensive training process, either online or onsite, during which participants were familiarized with the relevant knowledge, the functionality of the platform, and the trial process. Additionally, training documents were provided to participants for reference. Participants were asked to enter their personal information after they entered the platform. To further support participants, a video explaining the relevant knowledge, platform usage, and trial processes was displayed on the homepage.

**2.9 Trial design**

This study has been registered on clinicaltrials.gov (NCT 06031818). The study included a usability trial and a clinical trial. The usability trial aimed to evaluate the usability of the interpretable model. The clinical trial aimed to evaluate the effect of model explanation on clinical decisions. It followed a three-track approach: 1. without AI assistance, 2. with AI assistance, and 3. with AI plus explainability assistance. All the trials took place between December 2023 and March 2024. Twelve clinicians participated in the usability trial, while 10 clinicians took part in the clinical trial. These participants were not involved in the model establishment process and were unaware of the PHLF status of the patients.

**2.10 Usability trial**

The participants were asked to complete a usability trial before they started the clinical trial. During the trial, each participant was presented with 6 cases with varying predicted probability generated by the $VAE - MLP_{swe-cl}$ model. Specifically, there were 2 cases within each predicted probability interval: 0.1-0.3, 0.4-0.6, and 0.6-0.9, which was selected by an experienced radiologist with over 10 years of work experience in liver ultrasound (M.X.L). For each case, participants were provided with 2D-SWE images, relevant clinical variables, model prediction, corresponding predicted probability, and model explanation. Two forms of model explanation were provided: counterfactual explanations and LRP. Counterfactual explanations showcased a series of images, illustrating changes in the 2D-SWE image when transforming the classifier's decision from negative to positive across predicted probabilities from 0.1 to 0.9, at 0.1 intervals. LRP offered both global and local plots elucidating the contribution of different features.

**2.10.1 Case-level usability:**

After each case, clinicians were asked to complete questionnaires (see Supplementary Method A2) assessing the usability of counterfactual explanation and LRP. Evaluation indices for counterfactual explanations included understandability, classifier decision justification, visual quality, helpfulness, and confidence. Evaluation indices for LRP comprised understandability, classifier decision justification, helpfulness, and confidence. A five-point Likert scale was used for measurements of participants' agreement levels with corresponding statements. The ratings were: 1=strongly disagree, 2=disagree, 3=neutral, 4=agree, 5=strongly agree, with 1 indicating low usability and 5 indicating high usability.

**2.10.2 System-level usability**

After the clinicians finished all 6 cases, the System Causability Scale (SCS)[38] was used to measure the quality of system-level usability of explanations provided by the explanation system (see

Supplementary Method A3). Ten items were provided to evaluate the quality of explanations given by the proposed explainable system. The ratings were: 1=strongly disagree, 2=disagree, 3=neutral, 4=agree, 5=strongly agree, with 1 indicating low explanation quality and 5 indicating high explanation quality.

### 2.11. *In silico* clinical trial
### 2.11.1 Track 1: without AI assistance ($T_{No\_AI}$)

The participants were asked to complete this trial after they finished the usability trial. Each participant was presented with 80 cases from the test set. For each case, participants were provided with 2D-SWE images and relevant clinical variables. Participants first reviewed both image and clinical information before making predictions regarding the risk of PHLF (high or low) based on their clinical expertise. Additionally, participants were asked to indicate their confidence level in their predictions (0-100% with 10% intervals).

### 2.11.2 Track 2: with AI assistance ($T_{AI}$)

The participants were asked to complete this trial at least 2 weeks after they finished track 1 of the clinical trial. Each participant was presented with the same 80 cases as in track 1. For each case, except for 2D-SWE images and relevant clinical variables, the AI prediction and corresponding predicted probability were also provided. The participants were asked to make predictions regarding the risk of PHLF based on their clinical expertise and AI predictions. Similarly, participants were asked to indicate their confidence level in their predictions.

### 2.11.3 Track 3: with AI plus explanation assistance($T_{AI\_Exp}$)

The participants were asked to complete this trial at least 2 weeks after they finished track 2 of the clinical trial. Each participant was presented with the same 80 cases as in Track 1 and Track 2. For each case, except for 2D-SWE images, relevant clinical variables, the AI prediction, and corresponding predicted probability, two forms of model explanation including counterfactual explanations and LRP were also provided. The participants were asked to make predictions regarding the risk of PHLF based on their clinical expertise, AI predictions, and AI explanations. Similarly, participants were asked to indicate their confidence level in their predictions. Additionally, they were asked to indicate their satisfaction level for model explanations (0-100% with 10% intervals).

### 2.12 Statistical analysis

Statistical analyses were conducted using SPSS, version 20.0. Continuous variables in the training and test cohorts were compared using either the Student's t-test or the Mann-Whitney test, as appropriate. Categorical variables were compared using the χ2 test. A two-sided p-value of less than 0.05 indicated a significant difference. In the training cohort, univariate logistic analysis was utilized to identify significant predictors associated with symptomatic PHLF. These predictors were used in a stepwise multivariate logistic regression to identify independent factors for symptomatic PHLF. A clinical model was then developed using logistic regression based on these factors. The models' performance was compared using receiver operating characteristic curves (ROC) and the DeLong test. Thresholds for each model were determined using the highest Youden index in the training cohort. Patient-level performance metrics, such as accuracy, sensitivity, specificity, positive predictive value (PPV), and negative predictive value (NPV), were assessed and reported based on the median prediction from all images per patient. The paired samples t-test was used to compare the mean accuracy and mean confidence level between *in silico* clinical trial tracks 1 and 2, tracks 2 and 3. Additionally, it was utilized to compare the mean accuracy and mean satisfaction levels with model explanations between the senior and junior groups.

## 3. Results

### 3.1 Baseline characteristics

A total of 345 patients, with a median age of 55.0 years (IQR 47.0-64.0), were included in the study, comprising 305 males and 40 females. Among them, 265 patients were in the training set, while 80 were in the test set.

The baseline characteristics of the training and test cohorts were detailed in Supplementary Table A1. Symptomatic PHLF was experienced by 107 patients (31.0%), including 97 with PHLF grade B and 10 with PHLF grade C. Six patients with PHLF grade C died due to acute liver failure within 20 to 39 days post-surgery. The occurrence of symptomatic PHLF was comparable between the training cohort (80 patients, 30.1%) and the test cohort (27 patients, 33.8%), with no statistically significant difference observed ($p = 0.546$). Significant differences were found between the training and test cohorts in terms of prothrombin time (PT) level ($p = 0.002$), international normalized ratio (INR) level ($p < 0.001$), Milan criteria ($p = 0.008$), major hepatectomy ($p = 0.011$), and MELD score ($p = 0.012$).

### 3.2 Performance of the clinical model, VAE-MLP$_{swe}$ model and VAE-MLP$_{swe-cl}$ model

The results of univariate logistic regression showed significant differences in albumin (ALB), γ-glutamyl transpeptidase (GGT), PT, INR, ALBI score, Child-Pugh score, Child-Pugh grade, MELD score, cirrhosis, clinically significant portal hypertension (CSPH), tumor size, BCLC stage, total liver volume (TLV), resected liver volume (RLV), future liver remnant volume (FLR), and FLR ratio between symptomatic PHLF group and non-symptomatic PHLF group (all $p < 0.05$, Supplementary Table A2). Multivariate logistic regression analysis identified INR, CSPH, and FLR ratio as significant independent predictors of symptomatic PHLF (all $p < 0.05$, Supplementary Table A2). These three variables were included to establish the clinical model. The clinical model showed an AUC of 0.809 (95% CI 0.715-0.902) in five-fold cross-validation and an AUC of 0.684 (95% CI 0.571-0.784) in the test set (Fig. 3a).

In five-fold cross-validation, the VAE-MLP$_{swe}$ model showed a higher AUC (0.759, 95% CI: 0.666-0.853) than the traditional Densenet121 model (AUC 0.745, 95% CI: 0.662-0.829) and Resnet18 model (AUC 0.703, 95% CI: 0.618-0.788) (Supplementary Fig. A1). In the test set, the VAE-MLP$_{swe}$ model showed a higher AUC of 0.746 (95% CI: 0.637-0.837) than the traditional Densenet121 model (AUC 0.710, 95% CI: 0.598-0.806, $p = 0.261$) and Resnet18 model (AUC 0.715, 95% CI: 0.603-0.810, $p = 0.580$) without significant difference (Supplementary Fig. A1).

In five-fold cross-validation, the VAE-MLP$_{swe-cl}$ model showed a higher AUC (0.849, 95% CI: 0.783-0.915) than VAE-MLP$_{swe}$ (AUC 0.759, 95% CI: 0.666-0.853) (Fig. 3a). In the test set, the VAE-MLP$_{swe-cl}$ model showed a higher AUC of 0.828 (95% CI 0.727-0.903) than the VAE-MLP$_{swe}$ model (AUC 0.746, 95% CI: 0.637-0.837, $p = 0.064$), the clinical model ($p = 0.032$) and some clinical variables related to symptomatic PHLF prediction such as ALBI score (AUC 0.644, 95% CI: 0.529-0.748, $p = 0.016$), Child-Pugh score(0.612, 95% CI: 0.497-0.719, $p = 0.006$) and MELD score (AUC 0.529, 95% CI: 0.406-0.656, $p = 0.002$) (Fig. 3b).

The accuracy, sensitivity, specificity, PPV and NPV of clinical model, VAE-MLP$_{swe}$ and VAE-MLP$_{swe-cl}$ model were listed in Table 1.

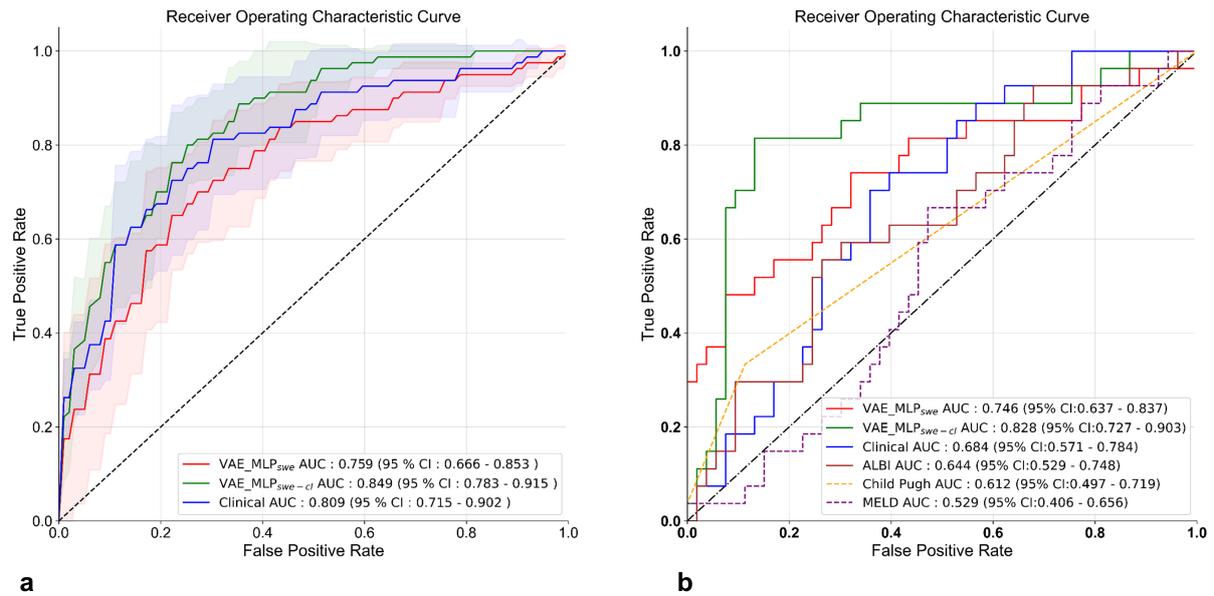

**Fig. 3 (a)** Receiver operating characteristic curves for VAE-MLP$_{swe}$ and VAE-MLP$_{swe-cl}$ model in five-fold cross-validation. **(b)** Receiver operating characteristic curves for VAE-MLP$_{swe}$, VAE-MLP$_{swe-cl}$ model, clinical model, and other clinical variables in the test cohort.

**Table 1** Five-fold cross-validation and test set results

| Model | AUC (95% CI) | Accuracy ±STD | Sensitivity ±STD | Specificity ±STD | PPV ±STD | NPV ±STD |
|---|---|---|---|---|---|---|
| **Five-fold cross-validation results** | | | | | | |
| VAE-MLP$_{swe}$ | 0.759 (0.666-0.853) | 0.728±0.059 | 0.700±0.133 | 0.741±0.095 | 0.548±0.080 | 0.856±0.051 |
| VAE-MLP$_{swe-cl}$ | 0.849 (0.783-0.915) | 0.762±0.050 | 0.750±0.105 | 0.768±0.065 | 0.587±0.064 | 0.878±0.042 |
| Clinical | 0.809 (0.715-0.902) | 0.739±0.051 | 0.713±0.170 | 0.751±0.035 | 0.549±0.057 | 0.865±0.078 |
| **Test set results** | | | | | | |
| VAE-MLP$_{swe}$ | 0.746 (0.637-0.837) | 0.688 | 0.704 | 0.679 | 0.527 | 0.818 |
| VAE-MLP$_{swe-cl}$ | 0.828 (0.727-0.903) | 0.813 | 0.815 | 0.811 | 0.686 | 0.896 |
| Clinical | 0.684 (0.571-0.784) | 0.650 | 0.550 | 0.698 | 0.484 | 0.755 |

AUC: area under the receiver operating characteristic curve; PPV: positive predictive value; NPV: negative predictive value

### 3.3 Qualitative and quantitative evaluation of counterfactual explanations

Counterfactual explanations were generated for the VAE-MLP$_{swe-cl}$ model with a predicted probability of 0.1-0.9. Four examples of qualitative evaluation of counterfactual explanations of images with different predicted probabilities were shown in Fig. 4a. More examples in GIF format can be found in Supplementary Fig. A2. The results showed that as the predicted probability increased, the color of the images generated by counterfactual analysis gradually shifted from blue to cyan, green, yellow, and finally red, which was consistent with the color changes in the color bar in the 2D-SWE images when the liver stiffness increased from low to high. Quantitative evaluation showed that the average liver stiffness increased as the predicted probability of counterfactual generated images increased (Fig. 4b), which was in line with clinicians' consensus that higher liver stiffness was associated with the occurrence of symptomatic PHLF in HCC[5]. Furthermore, Fig. 4b indicated that when the mean liver stiffness exceeded 11.5 kPa, there was a notable increase in the incidence of PHLF. This finding was consistent with the threshold of 11.5 kPa for diagnosis of liver cirrhosis recommended by the Aixplorer ultrasound system (SuperSonic Imagine S.A.) used in this trial.

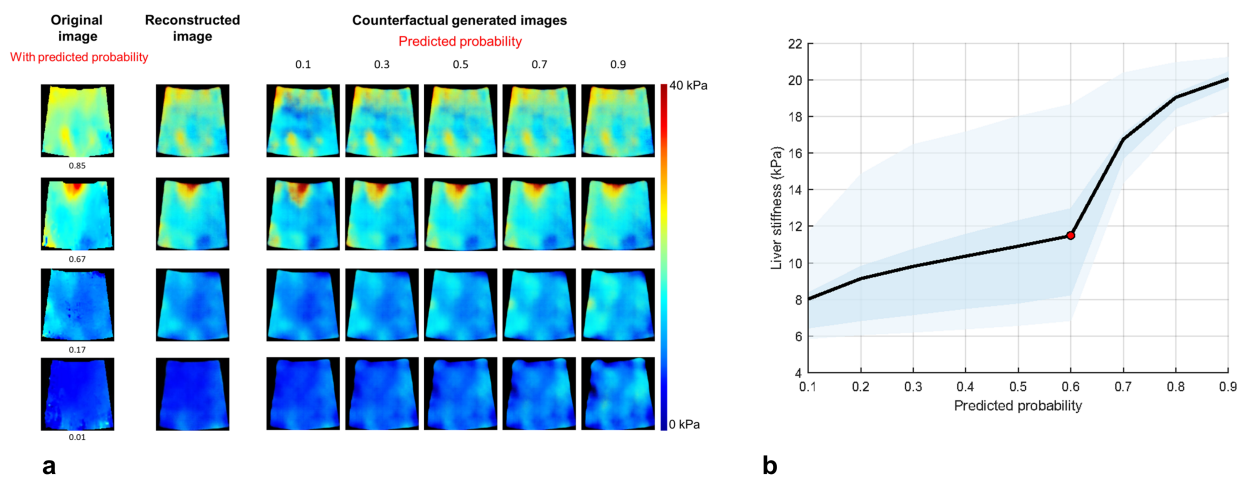

**Fig. 4** Qualitative and quantitative evaluation of counterfactual explanation. (a) Qualitative evaluation of counterfactual explanation. Each panel provides four examples where the first column shows original images with their classifier-predicted probabilities below. The second column reveals the corresponding reconstructed images. For these inputs, our model generates counterfactual images with predicted probabilities of 0.1, 0.3, 0.5, 0.7, and 0.9, displayed on the figure's right side. **(b)** Quantitative evaluation of counterfactual explanation. It shows the liver stiffness trend in counterfactual images with predicted probabilities ranging from 0.1 to 0.9. The average liver stiffness value for the test cohort is depicted by a black line, marked at 0.1 intervals. Dark blue transparent bands represent the 25%-75% confidence intervals, while light blue band indicates the 5%-25% and 75%-95% confidence intervals. The notable increase in the incidence of PHLF is marked by a red dot.

### 3.4 Qualitative evaluation of layer-wise relevance propagation

A Global LRP analysis of VAE-MLP$_{swe-cl}$ model identified 2D-SWE, FLR, and ALB as the most important features for symptomatic PHLF prediction. Other variables such as RLV, TBIL, TLV, MELD score, and PT, also contributed to the prediction. However, variables such as GGT, Child-Pugh score, BCLC staging, ALBI, tumor size, liver cirrhosis, Milan criteria, splenomegaly, CSPH, Child-Pugh grade, FLR ratio, INR, major hepatectomy, and ascites had very little contribution to the prediction (Fig. 5a). Figs. 5b and 5c show LRP local bar plots for two test cases. Fig. 5b shows a case without symptomatic PHLF that had been classified correctly by the model. The plot showed that FLR, SWE and ALBI contributed most to the negative prediction. Fig. 5c shows a case with symptomatic PHLF that has been classified incorrectly by the model. The plot showed that SWE, ALB, and FLR all contributed to the negative prediction. While we can see that although ALB and SWE were similar between these two cases, the FLR in the first case (1603.71ml) was larger than the second case (706.71ml). Thus, clinicians have the option to trust or question AI predictions based on the alignment of feature contributions from local LRP with their clinical expertise and understanding.

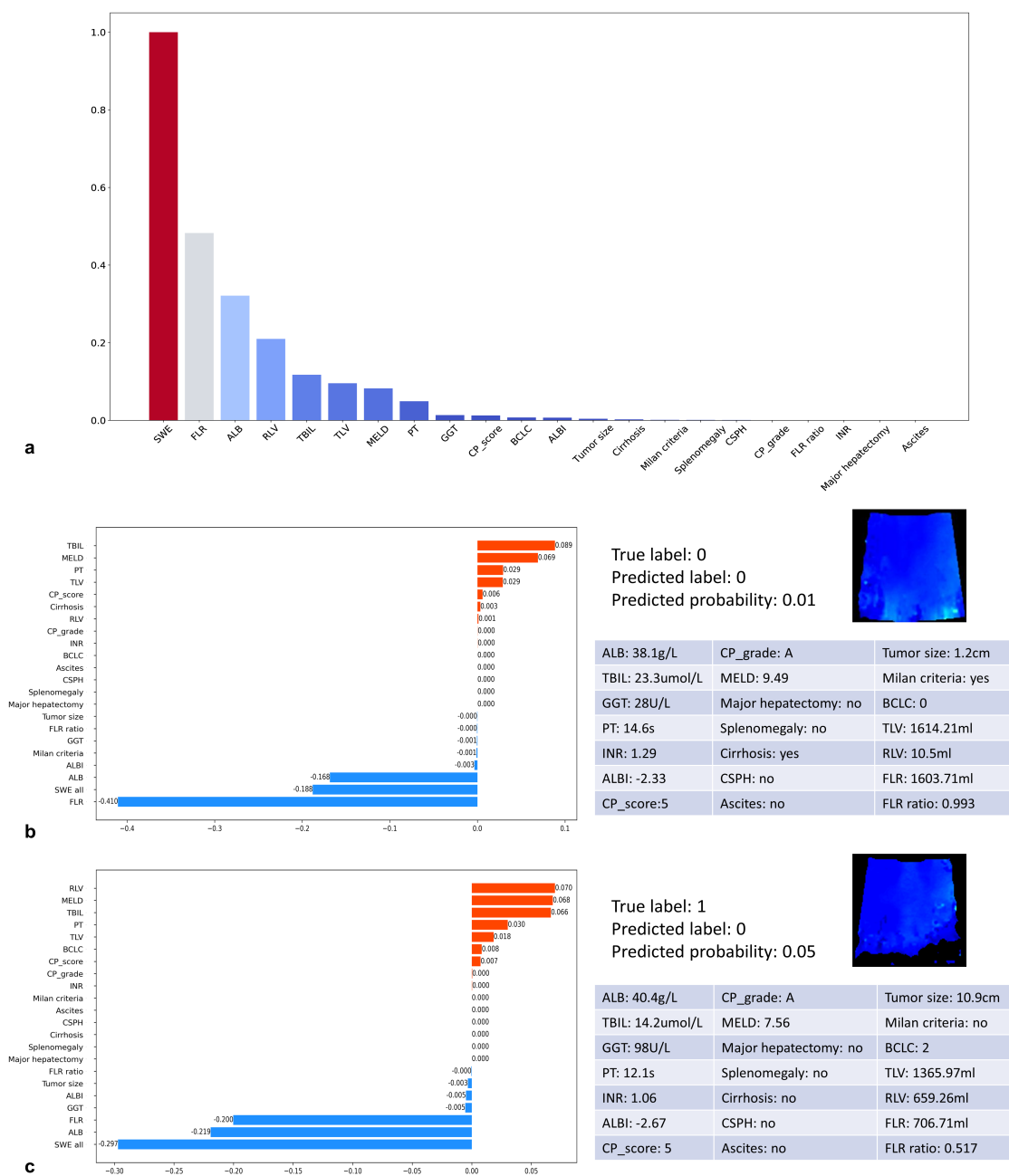

**Fig. 5 (a)** The global layer-wise relevance propagation shows different feature contributions in the whole test set. **(b)** The local layer-wise relevance bar plot shows the feature contributions for a case without symptomatic PHLF predicted correctly by the model. **(c)** The local layer-wise relevance bar plot shows the feature contributions for a case with symptomatic PHLF predicted incorrectly by the model. ALB, albumin; TBIL, total bilirubin; GGT, gamma-glutamyl transferase; PT, prothrombin time; INR, international normalized ratio; ALBI: Albumin-Bilirubin; CP_score: Child-Pugh score; CP_grade: Child-Pugh grade; MELD: a model for end-stage liver disease; CSPH: clinically significant portal hypertension; BCLC, Barcelona Clinic Liver Cancer; TLV: total liver volume; RLV: resected liver volume; FLR: future liver remnant volume; PHLF: post-hepatectomy liver failure.

### 3.5 Usability test results

A total of 12 international clinicians (9 from China, 2 from Belgium, and 1 from the Netherlands) specializing in general surgery (4 clinicians) or radiology (8 clinicians) were recruited to participate in the usability trial. Six clinicians had more than 5 years of work experience, while the remaining six had less than 5 years of experience.

The scores of case-level usability of counterfactual explanations and LRP were illustrated in Fig. 6a. For counterfactual explanations, the scores of usability corresponding to understandability, classifier's decision justification, visual quality, helpfulness, and confidence were 4.5±0.7, 4.2±0.9, 4.3±0.8, 4.4±0.6, and 4.3±0.8, respectively. The overall score for all five items was 4.4±0.8. For LRP, the scores of usability corresponding to understandability, classifier's decision justification, helpfulness, and confidence were 4.5±0.7, 4.2±0.8, 4.3±0.8, and 4.3±0.8, respectively. The overall score for all four items was 4.4±0.8.

Regarding the system-level System Causability Scale, the explanation system demonstrated good explanation quality with a mean score across ten items of 4.1±1.0, with the score of 4.2±0.9, 4.6±0.6, 4.1±0.5, 3.4±1.3, 4.5±0.7, 4.3±1.0, 3.7±1.1, 3.4±1.1, 4.3±0.6, and 4.2±1.0 for each item, respectively (Fig. 6b). Notably, scores were relatively low for items such as "No need for support" and "Quick to understand", indicating a higher cost of understanding the explanations compared to other aspects.

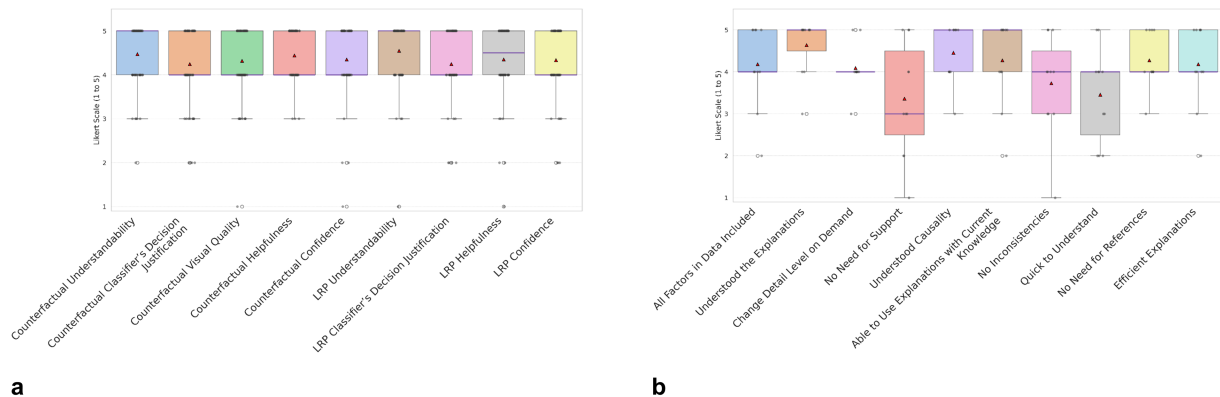

**a**                    **b**

**Fig. 6** Usability evaluation of counterfactual explanations and layerwise relevance propagation (LRP) on a case-level and system-level basis. **(a)** Case-level usability scores for counterfactual explanations and LRP, assessing understandability, classifier's decision justification, visual quality, helpfulness, and confidence, as rated by clinicians on a five-point Likert scale. The purple line indicates the median and the red triangle represents the mean values. **(b)** System-level usability evaluation using the System Causability Scale questionnaire responses from the clinicians, reflecting the overall quality of the explanation system.

### 3.6 *In silico* clinical trial results

A total of 10 international clinicians (8 from China, and 2 from Belgium) specializing in surgery or radiology were recruited to participate in the explainability trial. Among them, 4 clinicians specialized in surgery, and 6 specialized in radiology. Five clinicians belonged to the senior group, with more than 5 years of work experience, while the remaining 5 were in the junior group, with less than 3 years of experience.

#### 3.6.1 Accuracy results

We first investigated the impact of AI assistance ($T_{AI}$) on the clinicians' prediction accuracy compared to that without AI assistance ($T_{No\_AI}$). Out of 10 participants, we observed a performance improvement with AI support for 8 participants, and no change for 2 participants. The clinicians' mean accuracy increased from 67.2%±7.2% in $T_{No\_AI}$ to 71.1%±5.0% in $T_{AI}$, showing significant difference ($p = 0.004$, Fig. 7a). F1 score increased from $0.48 \pm 0.15$ in $T_{No\_AI}$ to $0.54 \pm 0.15$ in $T_{AI}$($p = 0.038$, Fig. 7b).

Then we investigated the impact of the model explanation ($T_{AI\_Exp}$) on the clinicians' prediction accuracy compared to $T_{AI}$. Out of 10 participants, we observed performance improvement with a model for 7 participants, performance decrease for 2 participants, and no change for one participant. Clinicians' mean accuracy increased from $71.1\% \pm 5.0\%$ in $T_{AI}$ to $73.6\% \pm 5.5\%$ in $T_{AI\_Exp}$ without significant difference($p = 0.117$, Fig. 7a). The mean F1 score increased from $0.54 \pm 0.15$ in $P_{AI}$ to $0.60 \pm 0.13$ in $T_{AI\_Exp}$($p = 0.060$, Fig. 7b).

To investigate the relationship between the clinicians' experience and benefit, we correlated their change in accuracy with their experience. The mean accuracy was $68.8\% \pm 4.2\%$ in the senior group and $63.8\% \pm 8.6\%$ in the junior group in $T_{No\_AI}$ ($p = 0.036$). The mean accuracy increased by 2.2% ($p = 0.308$) in the senior group ($71.0\% \pm 5.0\%$) and by 7.5% ($p = 0.003$) in the junior group ($71.3\% \pm 4.9\%$) when AI assistance was provided in $T_{AI}$. The mean accuracy increased by 4.8% ($p = 0.021$) in the senior group ($75.8\% \pm 3.4\%$) and by 0.2% ($p = 0.901$) in the junior group ($71.5\% \pm 6.2\%$) compared with $T_{AI}$ when the model explanation was provided in $T_{AI\_Exp}$(Fig. 7c). The above results indicated that the senior group benefited more from AI explanation while the junior group benefited more from only AI prediction.

The radar plots show that accuracy, negative predictive value (NPV), positive predictive value (PPV), recall, and specificity increased from $T_{No\_AI}$ to $T_{AI}$ to $T_{AI\_Exp}$. However, none of these indexes improved diagnostic metrics beyond AI alone (Fig. 7d).

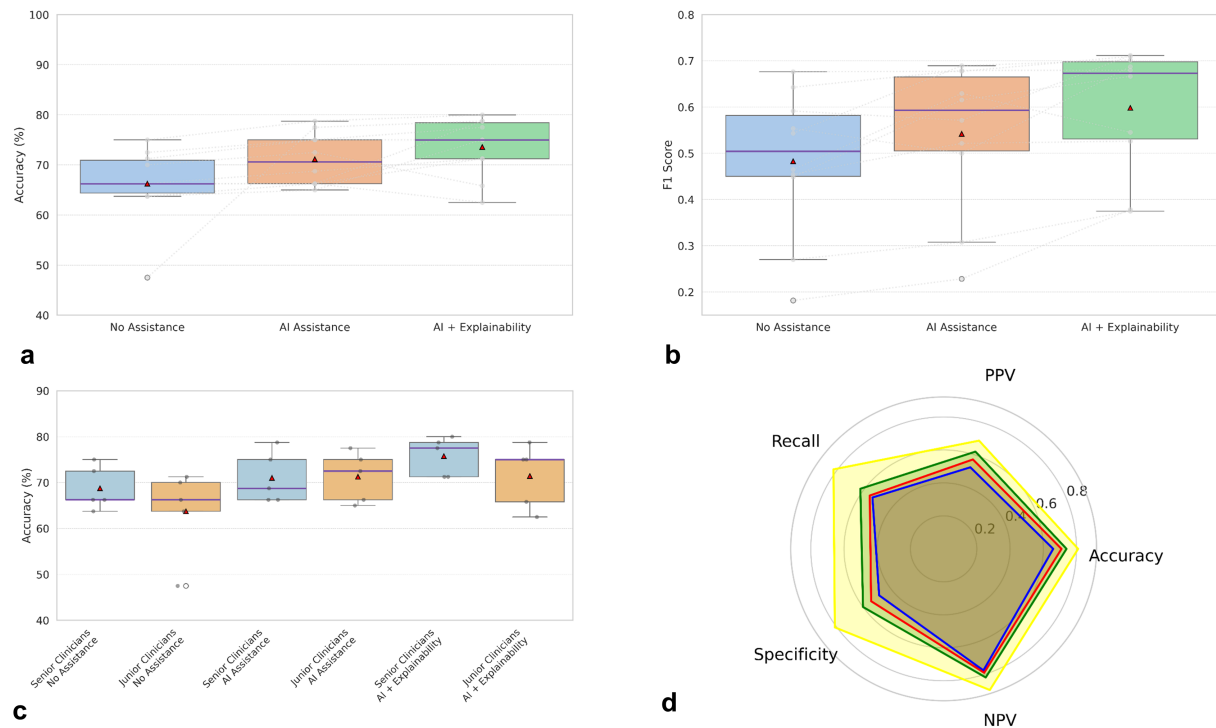

**Fig. 7** Comparative analysis of quantitative performance across different tracks of the *in silico* clinical trial. **(a)** Accuracy among clinicians without AI assistance, with AI assistance, and AI plus explainability. **(b)** F1 score across the three tracks. **(c)** Accuracy correlated with clinicians' level of experience in each track. **(d)** A radar plot showing the results of the *in silico* clinical trial without AI assistance (blue), with AI assistance (red), AI plus explainability assistance (green), and the AI model alone (yellow), comparing their impact on accuracy, NPV, PPV, recall, and specificity.

### 3.6.2 Confidence results

We first investigated the influence of AI assistance ($T_{AI}$) on clinicians' confidence compared to $T_{No\_AI}$. Out of 10 participants, we observed an increase in confidence with AI support for 7 participants and a decrease for 3 participants. The clinicians' mean confidence was 71.6%±12.8% in $T_{No\_AI}$ and 77.9%±11.7% in $T_{AI}$, showing a significant difference ($p < 0.001$, Fig. 8a).

Then we investigated the influence of model explanation ($T_{AI\_Exp}$) on clinicians' confidence compared to $T_{AI}$. Out of 10 participants, we observed an increase in confidence with model explanation for 7 participants and a decrease in 3 participants. The clinicians' mean confidence further increased to 79.2%±11.2% in $T_{AI\_Exp}$, showing a significant difference ($p = 0.002$, Fig. 8a).

We further investigated the influence of confidence on the clinicians' accuracy across the 3 tracks. The mean accuracy showed a trend of increase as the confidence level increased in all tracks (Fig. 8b). $T_{AI\_Exp}$ showed higher accuracy than $T_{No\_AI}$ and $T_{AI}$ while the confidence level ranged from 20% to 100% (Fig. 8b). After excluding cases where clinicians had the lowest 10% confidence level, the mean accuracy in all 3 tracks increased, showing mean accuracies of 70.0%±6.7%, 77.0%±4.2%, and 77.2%±6.8%, respectively (Fig. 8c).

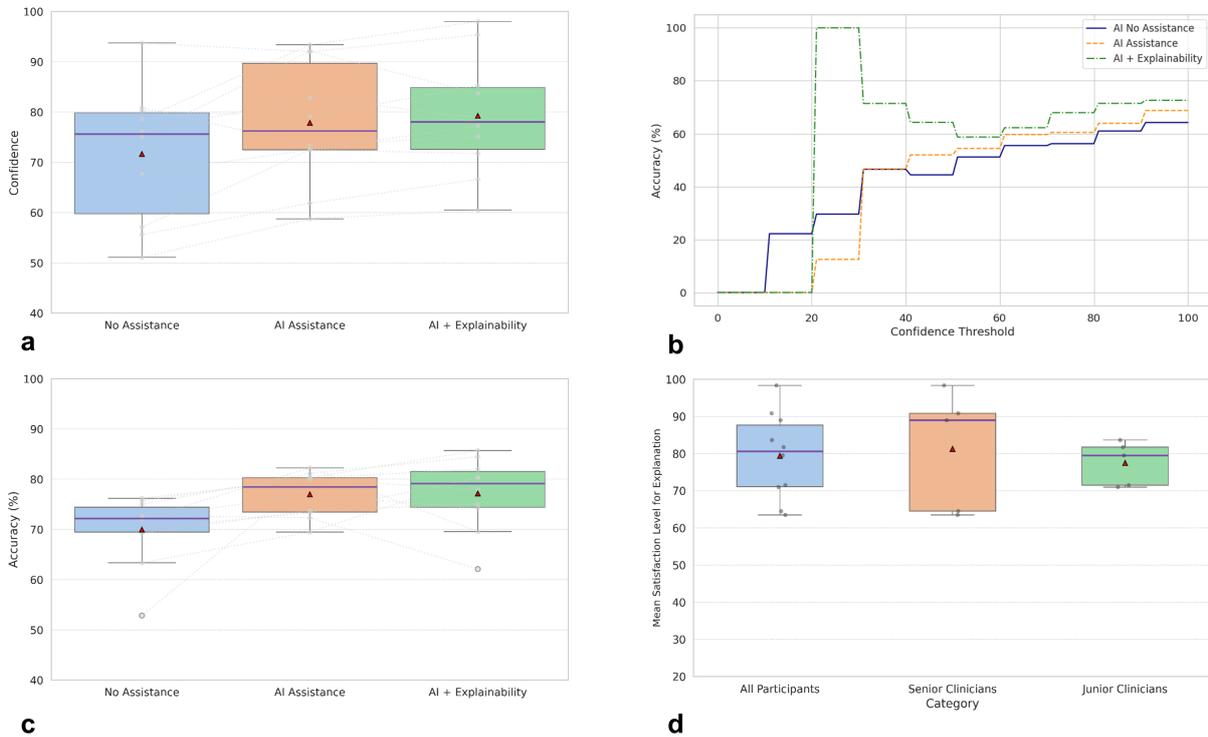

**Fig. 8** Influence of AI assistance and explainability on clinician confidence, accuracy, and satisfaction level. **(a)** Clinician confidence levels without AI assistance, with AI assistance, and with AI plus explainability assistance. The purple line represents the median, and the red triangle indicates the mean. **(b)** Accuracy relative to various confidence thresholds for all predictions across all participants. **(c)** Accuracy excluding predictions made with the lowest 10% confidence in three tracks. **(d)** Mean satisfaction levels with model explanations among all participants and between senior and junior groups in clinical trial with AI plus explainability assistance.

### 3.6.3 Results of satisfaction of explanation

We investigated the satisfaction level of clinicians with the model explanation in $T_{AI\_Exp}$. The mean satisfaction level in all participants was 79.4%±11.0% (Fig. 8d). The mean satisfaction level was 77.5%±5.3% in the junior group and 82.8%±15.7% in the senior group, showing a significant difference ($p < 0.001$, Fig. 8d).

## 4. Discussion

We developed an interpretable VAE-MLP model that combined 2D-SWE images and clinical variables for preoperative PHLF prediction in HCC. Our model explained the decision-making process by gradually perturbing the input image into counterfactual images to demonstrate "what-if" scenarios, and by displaying the contribution weights of both the images and clinical variables using the LRP method. The usability and efficiency of the model explanation were confirmed by a proposed evaluation framework including qualitative and quantitative assessments, usability evaluations, and an *in silico* clinical trial.

The proposed VAE-MLP model seamlessly integrates medical images and clinical variables for diagnosis and prediction, incorporating counterfactuals and LRP to provide insights into the model's decision-making mechanism. This approach holds the potential for generalization to other clinical diagnostic or predictive tasks requiring transparent model explanations. We perturbed the latent space of the VAE model to generate counterfactuals with increasing and decreasing probabilities for interpretability. Cohen et al. [48] used simple autoencoders and latent shifts to generate counterfactuals for chest X-ray classification. The classifiers were trained independent of the latent space representation of the autoencoders. Singla et al. [49] employed generative adversarial networks (GANs) to generate counterfactuals for multi-label classification for chest X-ray images. In comparison to the aforementioned study, we used the latent space of the VAE to perform a classification task, which was a simpler approach to counterfactual generation. In addition, clinical variables were concatenated to the latent representation for better prediction. The integration of image features and clinical variables is essential for the accurate diagnosis and prediction of many clinical issues. In our study, we found that the VAE-MLP$_{swe-cl}$ model outperformed the VAE-MLP$_{swe}$ model, consistent with previous findings[50,51]. We employed LRP analysis to determine the feature attributions of both medical images and clinical variables, empowering doctors to scrutinize or trust the model's predictions based on these insights. In the future, there is a need to extend the proposed VAE-MLP model with two complementary streams of explanation—counterfactuals and LRP—to other medical AI applications.

We are one of the first studies to propose a methodological framework for the explainability evaluation, including qualitative and quantitative assessments, usability evaluations, and an three-track *in silico* clinical trial that included evaluations without AI assistance, with AI assistance, and with AI plus explainability assistance. The model's explanation was first evaluated through quantitative and qualitative assessments using established biomarkers and existing clinical knowledge. The qualitative evaluation highlighted the clinical relevance of changes observed in counterfactual images[52]. We used liver stiffness measurement for quantitative verification of the quality of counterfactual explanations. Global LRP analysis of VAE-MLP$_{swe-cl}$ model identified 2D-SWE, FLR, and ALB as the most important features for symptomatic PHLF prediction. PHLF is closely associated with both the quality and quantity of the remaining liver tissue[53]. 2D-SWE provided crucial information regarding the quality or function of the remaining liver tissue[10], while FLR, representing the volume of the remnant liver, provided information on the quantity of liver tissue that remained[54]. Additionally, ALB was identified as a critical feature in predicting PHLF due to its association with liver function[55]. Our results demonstrated that LRP analysis provided high interpretability for both images and clinical variables and consistency with previous studies. Quantitative and qualitative evaluation verifies the reliability and correctness of explanations before they can be used in a clinical setting.

Additionally, conducting expert evaluations is critical for assessing the usability of these explanations from the clinicians' perspective. Our study marked one of the first attempts to evaluate the usability of the interpretable model before delving into its clinical efficacy [40,56]. The usability of our model explanation underwent clinical assessment at both the case-level and system-level. The usability scores for both counterfactual explanations and LRP indicated a high level of user satisfaction across various dimensions such as understandability, classifier's decision justification, helpfulness, and confidence[49]. This suggested that both counterfactual explanations and LRP were effective in providing explanations that users found comprehensible, justifiable, and supportive of their decision-making processes. The overall quality of the proposed model explanation system was assessed using the SCS, showing generally good explanation quality across ten items. However, it is important to note that certain aspects, such as the need for support to understand the explanations and the perceived learning curve for understanding the explanations, received relatively lower scores. This suggests that there may be a higher cognitive cost associated with understanding the explanations, which could impact users' overall experience and adoption of the system. More future research should be done to establish the explanation model, which is technically accurate and informative but, in the meantime, easily understandable and accessible to users without imposing a significant cognitive burden.

Furthermore, it is important to investigate the impact of introducing AI and AI explanations in a collaborative setting involving clinicians. Leveraging the outcomes of in silico trials could facilitate expedited approval processes by regulatory bodies or certification authorities. To ensure the integrity of our explainability assessment, we pre-registered the *in silico* clinical trial to avoid post hoc hypothesizing based on the results. The results from the *in silico* clinical trial highlighted the significant impact of AI assistance and model explanation on clinicians' prediction accuracy. However, despite these advancements, they did not surpass the diagnostic performance achieved solely with the AI model. Further research is warranted to assess and enhance the interaction between clinicians and AI systems in real clinical environments. Notably, the senior group benefited more from AI explanation, while the junior group showed greater benefits from AI prediction, which was consistent with the research of Chanda et al[40], highlighting the potential differential impact based on clinicians' experience levels. Future research should explore how experienced and inexperienced clinicians interact with explainable AI systems. We observed that clinicians' confidence increased with AI support and that AI explanation enhanced this effect even further. Interestingly, we observed a consistent trend of increased accuracy with higher confidence levels. Furthermore, upon excluding cases where clinicians had the lowest 10% confidence level, the accuracy showed notable improvements. These results underscored the role of confidence in clinicians' accuracy and highlighted the potential of AI and model explanation assistance in boosting both confidence and accuracy levels. The results indicated overall high satisfaction with explanations, with the senior group showing even higher satisfaction compared to the junior group. This difference in satisfaction levels could be attributed to the fact that experienced clinicians likely have a deeper appreciation or understanding of model explanation, as they possess a clear understanding of the contribution of each variable in the context of their expertise.

There were a few limitations in the study. Firstly, only a few clinicians were enrolled for the usability trial and *in silico* clinical trial, future trials of higher sample size and external validation will be needed to strengthen the evidence base for the effectiveness of model explanation in clinical practice. Secondly, the usability test indicated a higher cost of understanding the explanations compared to other aspects. Hence, there is still room for improvement in explaining complex AI algorithms to non-expert users. Future work should focus on incorporating interactive visualizations, simplified language, and user-friendly interfaces that can enhance explanation understanding and facilitate trust in AI-based tools. Finally, during the development of the explainable AI model, only a few clinicians were involved. Because of the different roles and knowledge of AI developers and end users, there is a need to conduct formative user research to understand user needs and domains during the design and development of explainable AI models.

In conclusion, we have developed an interpretable VAE-MLP model for accurate prediction of PHLF in HCC. The use of counterfactual generation and LRP provided interpretability for the model. Besides, we proposed a comprehensive framework for explainability evaluation of the model. The quantitative and qualitative evaluation of counterfactuals and qualitative evaluation of LRP analysis correlated with clinical knowledge, demonstrating the validity of the explanation. Our explanation framework showed good usability at both the case-level and system-level and the results highlighted the impact of AI and AI explanation on clinicians' prediction accuracy, confidence, and satisfaction. The VAE-MLP model and the associated explanation evaluation framework hold promise for broader application in medical contexts, enhancing the interpretability of medical AI.

## Author contributions statements

Xian Zhong and Zohaib Salahuddin: Conceptualization, Methodology, Formal analysis, Data curation, Writing - original draft, Project administration. Yi Chen: Methodology. H.C.Woodruff: Supervision, Writing - review and editing. Haiyi Long, Jianyun Peng, Nuwan Udawatte, Roberto Casale, Ayoub Mokhtari, Xiaoer Zhang, Jiayao Huang, Qingyu Wu, Li Tan, Lili Chen and Dongming Li: Data curation. Xiaoyan Xie: Conceptualization, Investigation, Project administration, Resources, Supervision, Writing – review & editing. Manxia Lin: Conceptualization, Data curation, Funding acquisition, Project administration, Supervision, Writing – review & editing. Philippe Lambin: Conceptualization, Data curation, Funding acquisition, Project administration, Supervision, Writing – review & editing.

## Acknowledgements

Authors acknowledge financial support from ERC advanced grant (ERC-ADG-2015 n° 694812 - Hypoximmuno) and ERC-2020-PoC: 957565-AUTO.DISTINCT. Authors also acknowledge financial support from the European Union's Horizon research and innovation programme under grant agreement: ImmunoSABR n° 733008, MSCA-ITN-PREDICT n° 766276, CHAIMELEON n° 952172, EuCanImage n° 952103, IMI-OPTIMA n° 101034347, AIDAVA (HORIZON-HLTH-2021-TOOL-06) n°101057062, REALM (HORIZON-HLTH-2022-TOOL-11) n° 101095435 and EUCAIM (DIGITAL-2022-CLOUD-AI-02) n°101100633. This work was supported by the Major Research plan of the National Natural Science Foundation of China (92059201); Natural Science Foundation of Guangdong Province, China (2023A1515012464); Guangzhou basic and applied basic research foundation (SL2023A04J02221), Natural Science Foundation of Guangdong Province, China (2022A1515011716).

## Disclosures

PL disclosures from the last 36 months within and outside the submitted work: grants/sponsored research agreements from Radiomics SA, Convert Pharmaceuticals and LivingMed Biotech. PL received a presenter fee (in cash or in kind) and/or reimbursement of travel costs/consultancy fee (in cash or in kind) from Radiomics SA, BHV. PL has minority shares in the companies Radiomics SA, Convert pharmaceuticals, Comunicare, LivingMed Biotech and Bactam. PL is co-inventor of two issued patents with royalties on radiomics (PCT/NL2014/050248 and PCT/NL2014/050728), licensed to Radiomics SA; one issued patent on mtDNA (PCT/EP2014/059089), licensed to ptTheragnostic/DNAmito; one non-issued patent on LSRT (PCT/ P126537PC00, US: 17802766), licensed to Varian; three non-patented inventions (softwares) licensed to ptTheragnostic/DNAmito, Radiomics SA and Health Innovation Ventures and two non-issued, non-licensed patents on Deep Learning-Radiomics (N2024482, N2024889). He confirms that none of the above entities were involved in the preparation of this paper. HW has minority shares in the company Radiomics SA,

# Supplementary

Table A1 Baseline characteristics of enrolled patients

| Characteristic | All patients (n = 345) | Training cohort (n=265) | Test cohort (n=80) | p value |
|---|---|---|---|---|
| Age (year) | 55.0 (47.0-64.0) | 55.0 (47.0-64.0) | 54.0 (49.0-66.8) | 0.354 |
| Sex (Male/Female) | 305/40 | 238/27 | 67/13 | 0.138 |
| Underlying liver disease (HBV/HCV/Coinfection of HBV and HCV/Unknown) | 324/7/6/8 | 249/5/5/6 | 75/2/1/2 | 0.965 |
| TBIL (umol/L) | 13.8 (10.7-17.3) | 13.6 (10.6-16.9) | 15.0 (11.5-17.9) | 0.081 |
| ALB (g/L) | 38.3 (36.2-41.0) | 38.3 (36.2-41.0) | 38.8 (36.2-41.2) | 0.942 |
| CREA (umol/L) | 79.0 (68.0-87.0) | 79.0 (68.0-87.5) | 80.0 (68.3-87.0) | 0.940 |
| ALT (U/L) | 31.0 (21.0-43.5) | 32.0 (20.0-43.0) | 31.0 (22.0-52.8) | 0.805 |
| AST (U/L) | 35.0 (25.0-50.0) | 35.0 (26.0-50.0) | 36.0 (23.0-50.7) | 0.812 |
| GGT (U/L) | 55.0 (34.0-98.5) | 59.0 (36.0-103.0) | 50.0 (30.3-85.8) | 0.055 |
| PT (s) | 11.9 (11.3-12.6) | 11.8 (11.2-12.4) | 12.2 (11.7-12.8) | 0.002 |
| INR | 1.02 (0.97-1.07) | 1.01 (0.96-1.06) | 1.05 (1.00-1.09) | <0.001 |
| AFP (U/L) | 23.1 (4.4-516.1) | 21.2 (4.5-527.4) | 49.6 (4.1-476.3) | 0.829 |
| ALBI | -2.52 [(-2.72)-(-2.34)] | -2.53 [-(2.73)-(-2.34)] | -2.54 [-(2.67)-(-2.33)] | 0.863 |
| ALBI grade (1/2) | 137/208 | 104/161 | 33/47 | 0.748 |
| Child-Pugh score (5/6/7) | 276/53/16 | 211/39/15 | 65/14/1 | 0.234 |
| Child-Pugh grade (A/B) | 329/16 | 250/15 | 79/1 | 0.100 |
| MELD | 4.8 (2.9-6.3) | 4.6 (2.6-6.2) | 5.4 (3.9-7.3) | 0.012 |
| Cirrhosis (Yes/No) | 120/225 | 90/175 | 30/50 | 0.560 |
| CSPH (Yes/No) | 39/306 | 29/236 | 10/70 | 0.700 |
| Splenomegaly (Yes/No) | 101/244 | 83/182 | 18/62 | 0.129 |
| Ascite (Yes/No) | 22/323 | 19/246 | 3/77 | 0.273 |
| Tumor size (cm) | 5.4 (3.5-8.3) | 5.7 (3.6-8.4) | 4.5 (3.0-7.5) | 0.107 |
| BCLC stage (0/A/B/C) | 19/222/62/42 | 14/163/49/39 | 5/59/13/3 | 0.051 |
| Milan criteria (Yes/No) | 95/170 | 95/170 | 42/38 | 0.008 |
| Major hepatectomy (Yes/No) | 114/231 | 97/168 | 17/63 | 0.011 |
| TLV (ml) | 1242.4 (1083.4-1528.2) | 1242.4 (1086.1-1531.2) | 1230.6 (1070.5-1526.0) | 0.707 |
| **RLV (ml)** | 428.0 (234.7-687.5) | 433.1 (234.7-704.9) | 366.1 (226.1-633.3) | 0.229 |
| **FLR** | 788.6 (643.5-963.1) | 788.6 (631.6-957.7) | 787.2 (689.9-1003.6) | 0.465 |
| **FLR ratio** | 0.67 (0.50-0.80) | 0.66 (0.48-0.79) | 0.69 (0.54-0.80) | 0.186 |
| **Symptomatic PHLF (Yes/No)** | 107/238 | 80/185 | 27/53 | 0.546 |

Continuous variables are expressed in median (P25–P75). Categorical variable are expressed in counts. TBIL, total bilirubin; ALB, albumin; CREA, creatinine; ALT, alanine aminotransferase; AST, aspartate transaminase; GGT, gamma-glutamyl transferase; PT, prothrombin time; INR, international normalized ratio; AFP, alpha-fetoprotein; ALBI: Albumin-Bilirubin; MELD: model for end-stage liver disease; CSPH: Clinically significant portal hypertension;  BCLC, Barcelona Clinic Liver Cancer;  TLV: total liver volume; RLV: resected liver volume; FLR: future liver remnant volume; PHLF: post-hepatectomy liver failure.

**Table A2  Influencing clinical factors of symptomatic PHLF**

| Variables | Univariate analysis | p value | Multivariate analysis | |
|---|---|---|---|---|
| | OR (95% CI) | | OR (95% CI) | p value |
| Sex, female vs. male | 0.791 (0.320-1.953) | 0.791 | - | - |
| Age (years) | 1.003 (0.981-1.025) | 0.811 | - | - |
| TBIL (umol/L) | 1.033 (0.998-1.069) | 0.062 | - | - |
| ALB (g/L) | 0.900 (0.838-0.967) | 0.004 | - | - |
| CREA (umol/L) | 0.997 (0.988-1.006) | 0.562 | - | - |
| ALT (U/L) | 1.001 (0.998-1.005) | 0.446 | - | - |
| AST (U/L) | 1.002 (0.999-1.006) | 0.234 | - | - |
| GGT (U/L) | 1.003 (1.000-1.005) | 0.024 | - | - |
| PT (s) | 1.343 (1.058-1.706) | 0.015 | - | - |
| INR | 2461.350 (70.906-85440.280) | < 0.001 | 2424.484 (49.342-119130.427) | < 0.001 |
| AFP (U/L) | 1.000(1.000-1.000) | 0.244 | - | - |
| ALBI score | 4.533 (1.947-10.557) | < 0.001 | - | - |
| Child-Pugh score | 2.031 (1.292-3.193) | 0.002 | - | - |
| Child-Pugh grade, B vs A | 3.782 (1.299-11.013) | 0.015 | - | - |
| MELD | 1.115(1.018-1.222) | 0.019 | - | - |
| Cirrhosis, yes vs. no | 2.499 (1.450-4.307) | 0.001 | - | - |
| CSPH, yes vs. no | 3.308 (1.507-7.260) | 0.003 | 4.670 (0.001-0.023) | 0.001 |
| Splenomegaly, yes vs. no | 1.618 (0.931-2.811) | 0.088 | - | - |
| Ascite, yes vs. no | 2.218 (0.865-5.689) | 0.097 | - | - |
| Tumor size (cm) | 1.177 (1.090-1.272) | < 0.001 | - | - |
| BCLC stage | 1.536 (1.113-2.118) | 0.009 | - | - |
| Milan criteria, yes vs. no | 0.407 (0.223-0.742) | 0.003 | | |
| Major hepatectomy, yes vs. no | 2.819 (1.640-4.847) | < 0.001 | | |
| TLV (ml) | 1.001 (1.000-1.002) | 0.001 | - | - |
| RLV (ml) | 1.002 (1.001-1.003) | < 0.001 | - | - |
| FLR (ml) | 0.997 (0.996-0.998) | < 0.001 | - | - |
| FLR ratio | 0.009 (0.002-0.039) | < 0.001 | 0.004 (0.001-0.023) | < 0.001 |

TBIL, total bilirubin; ALB, albumin; CREA, creatinine; ALT, alanine aminotransferase; AST, aspartate transaminase; GGT, gamma-glutamyl transferas; PT, prothrombin time; INR, international normalized ratio; AFP, alpha-fetoprotein; ALBI: Albumin-Bilirubin; MELD: Model for end-stage liver disease; CSPH: Clinically significant portal hypertension; BCLC, Barcelona Clinic Liver Cancer; TLV: total liver volume; RLV: resected liver volume; FLR: future liver remnant volume

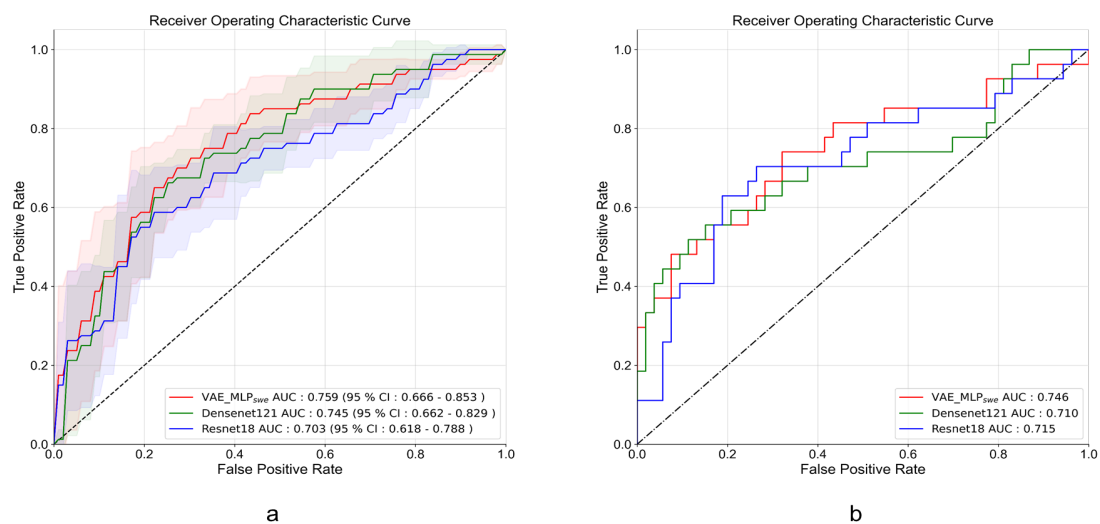

**Fig. A1** (a) Receiver operating characteristic curves for VAE-MLP$_{swe}$ model, Densenet121 model and Resnet18 model in five-fold cross-validation. (b) Receiver operating characteristic curves for VAE-MLP$_{swe}$ model, Densenet121 model and Resnet18 model in the test cohorts.

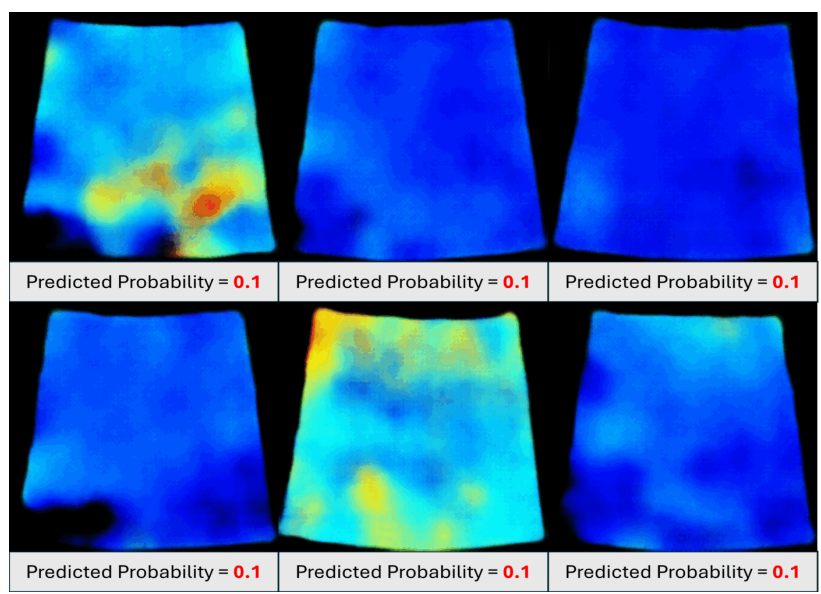

**Fig. A2** Graphics Interchange Format (GIF) showing counterfactual explanations corresponding to different probabilities for two-dimensional shear wave elastography (2D-SWE) image. GIF is available at the following link: https://drive.google.com/file/d/1jV6kTpH1-xoNVCZ6kVcNaC_8-chH63YN/view?usp=sharing

**Method A1**

Dataset and clinical data collection

The diagnosis of HCC was based on American Association for the study of liver diseases (AASLD) clinical practice guidelines for HCC ( Edition 2018), and the staging of HCC was based on Barcelona Clinic Liver Cancer (BCLC) staging (Edition 2018) and China Liver Cancer (CNLC) staging (Edition 2019). The inclusion criteria were as follows: 1) patients with treatment-naive and resectable HCC (defined as CNLC stage I to IIIa); 2) performance status Eastern Cooperative Oncology Group (PS) score 0–1. The exclusion criteria were as follows: 1) liver resection was not performed; 2) pathological diagnosis of non-HCC; 3) failure in liver stiffness measurement defined as the elastography color map was less than 75% filled or interquartile range (IQR)/median >30%; 4) immune-active chronic hepatitis indicated by an elevation of alanine aminotransferase (ALT) levels ≥2×upper limit of normal (ULN); 5) obstructive jaundice or dilated intrahepatic bile ducts with a diameter of >3 mm; 6) hypoalbuminemia, hyperbilirubinemia, or coagulopathy not related to the liver.

For each patient, 3-10 2D-SWE images of liver parenchyma were obtained using Aixplorer ultrasound system (SuperSonic Imagine S.A.) with a convex broadband probe (SC6-1, 1–6 MHz). The depth of 2D-SWE measurement was set at 4–6 cm, and the scale was set as 40 kPa. The sampling frame was 4×3 cm in size. All images were stored in the Digital Imaging and Communications in Medicine (DICOM) format. 2D-SWE images display color-coded tissue stiffness maps of liver parenchyma, with red representing a solid tissue (higher stiffness) and blue representing a soft tissue (lower stiffness).

Preoperative patient characteristics, laboratory data, and radiological data including upper abdominal computed tomography (CT) and magnetic resonance imaging (MRI) were collected within one week before surgery. Clinically significant portal hypertension (CSPH) was defined as the presence of esophageal varices (by CT/MR) and/or platelet count<$100×10^9$/L in association with splenomegaly. Splenomegaly was defined as the longest diameter of the spleen greater than 12 cm measured on coronal and axial CT/MRI images in the portal venous phase. Total liver volume (TLV), resected liver volume (RLV), and future liver remnant volume (FLR) were assessed based on 3-dimensional reconstruction and simulation of the surgical resection plan on preoperative CT or MRI imaging based on IQQA-Liver system (EDDA Technology). FLR ratio was defined as liver remnant volume/total liver volume to represent the percentage of the remnant liver after resection. The Child-Pugh score, albumin-bilirubin (ALBI) score, and end-stage liver disease (MELD) score were calculated according to formulas presented below:

(1) The Child-Pugh score was based on the total bilirubin, albumin, prothrombin time, and the clinical findings of encephalopathy and ascites. It was graded as 5–6 points for Child-Pugh grade A; 7–9 points for Child-Pugh grade B; and 10–15 points for Child-Pugh grade C.

(2) The following formula determined the ALBI score: (log10 bilirubin μmol/L × 0.66) + (−0.085 × albumin g/L). The ALBI score was graded as: score ≤−2.60 as ALBI grade 1; −2.60< score ≤−1.39 as ALBI grade 2; and score >−1.39 as ALBI grade 3.

(3) The MELD score was calculated according to the formula: 3.8 × loge (bilirubin (mg/dl)) + 11.2 × loge (INR) + 9.6 × loge (creatinine (mg/dl)) + 6.4 × (etiology: 0 if cholestatic or alcoholic, 1 otherwise).

**Method A2**

*Questionnaires on case-level usability*

    I.    **Questionnaire for counterfactual explanation**

        1. Understandability: I understand how the AI system made the above assessment for PHLF.

        2. Classifier's decision justification: The changes in the SWE image are related to PHLF.

        3. Visual quality: The generated counterfactual images look like SWE images.

        4. Helpfulness: The explanation helped me understand the assessment made by the AI system.

        5. Confidence: I feel more confident about the model with the explanation.

    II.    **Questionnaire for layerwise relevance propagation**

        1. Understandability: I understand which features influence the prediction and how they influence it.

        2. Classifier's decision justification: The feature's contribution is reasonably related to PHLF.

        3. Helpfulness: The explanation helped me understand the assessment made by the AI system.

        4. Confidence: I feel more confident about the model with the explanation.

**Method A3**

System Causability Scale

1. I found that the data included all relevant known causal factors with sufficient precision and granularity.
2. I understood the explanations within the context of my work.
3. I could change the level of detail on demand.
4. I did not need support to understand the explanations.
5. I found the explanations helped me to understand causality.
6. I was able to use the explanations with my knowledge base.
7. I did not find inconsistencies between explanations.
8. I think that most people would learn to understand the explanations very quickly.
9. I did not need more references in the explanations: e.g., medical guidelines, and regulations.
10. I received the explanations in a timely and efficient manner